\documentclass[letterpaper, 10 pt, conference]{ieeeconf}  

\IEEEoverridecommandlockouts                              

\usepackage{amsthm}

\usepackage{cite}
\usepackage{amsmath,amssymb,amsfonts}
\usepackage{algorithmic}
\usepackage{graphicx}
\usepackage{textcomp}
\usepackage{xcolor}

\usepackage{amsmath}
\usepackage{amsthm}
\usepackage{algorithm}
\usepackage{algorithmic}
\usepackage{multirow}
\usepackage{amssymb}

\newtheorem{theorem}{Theorem}
\newtheorem{definition}{Definition}
\newtheorem{lemma}{Lemma}

\title{\LARGE \bf
Phasic Diversity Optimization for Population-Based Reinforcement Learning
}

\author{Jingcheng Jiang$^{1}$, Haiyin Piao$^{2}$, Yu Fu$^{1}$, Yihang Hao$^{3}$, Chuanlu Jiang$^{1}$, Ziqi Wei$^{4,*}$, Xin Yang$^{1,*}$
\thanks{$^{1}$J. Jiang, Y. Fu, C. Jiang and X. Yang are with Key Laboratory of Social Computing and Cognitive Intelligence, Dalian University of Technology, Dalian, China, {\tt\small\{jiangjingcheng, fuyuu, chuanlujiang\}@mail.dlut.edu.cn; xinyang@dlut.edu.cn}}%
\thanks{$^{2}$H. Piao is with School of Electronics and Information, Northwestern Polytechnical University, Xi'an, China, {\tt\small haiyinpiao@mail.nwpu.edu.cn}}
\thanks{$^{3}$Y. Hao is with Yangzhou Collaborative Innovation Research Institute Co., Ltd, Yangzhou, China, {\tt\small hyh19951114@gmail.com}}
\thanks{$^{4}$Z. Wei is with Institute of Automation, Chinese Academy of Sciences, Beijing, China, {\tt\small ziqi.wei@ia.ac.cn}}
\thanks{$^{*}$Corresponding author}
}

\begin{document}

\maketitle
\thispagestyle{empty}
\pagestyle{empty}

\begin{abstract}
Reviewing the previous work of diversity Reinforcement Learning, diversity is often obtained via an augmented loss function, which requires a balance between reward and diversity. Generally, diversity optimization algorithms use Multi-armed Bandits algorithms to select the coefficient in the pre-defined space. However, the dynamic distribution of reward signals for MABs or the conflict between quality and diversity limits the performance of these methods. We introduce the Phasic Diversity Optimization (PDO) algorithm, a Population-Based Training framework that separates reward and diversity training into distinct phases instead of optimizing a multi-objective function. In the auxiliary phase, agents with poor performance diversified via determinants will not replace the better agents in the archive. The decoupling of reward and diversity allows us to use an aggressive diversity optimization in the auxiliary phase without performance degradation. Furthermore, we construct a dogfight scenario for aerial agents to demonstrate the practicality of the PDO algorithm. We introduce two implementations of PDO archive and conduct tests in the newly proposed adversarial dogfight and MuJoCo simulations. The results show that our proposed algorithm achieves better performance than baselines.

\end{abstract}

\section{INTRODUCTION}

Deep Reinforcement Learning (RL) has been successfully applied to robotic grasping, obstacle avoidance, navigation, and automatic driving. The essence of RL is the exploitation of known knowledge and the exploration of unknown environments. Whereas exploitation refers to policy improvement from environmental interaction, exploration refers to the actions sampled from the conditional distribution. 

Recent studies have shown great interest in achieving precise aerial tracking and maneuver avoidance through RL. Agents need enough exploration to escape from sub-optimal policies. "Natural selection and survival of the fittest". The Population-Based Training (PBT,~\cite{pbt}) provides an elegant way to balance exploration and exploitation, that is, to train agents of a set of diverse behaviors asynchronously and accelerate training by copying the information of the elite in the population. However, the expensive cost of large-scale distributed training~\cite{pga-me} accounts for the limitation of the number of learners. On the other hand, the small-scale training or inappropriate exploitation frequency leads to rapid convergence to local optima. Therefore, maintaining the diversity of the small-scale population becomes popular, and modeling diversity in a unified and principled way is still an open issue in the field of RL~\cite{divauto,neuro}. For example, the diversity of most algorithms is obtained from the expectation of some distance (or divergence) between pairwise policies regardless of the differences in norms between different pairwise similarities.

The determinantal point process (DPP,~\cite{dpp,dvd}) provides a solution for efficient diversity modeling and prevents the clustering phenomenon by a high-order joint optimization. The diversity is derived from a determinant of a positive semidefinite (PSD) matrix that defines a global measure of similarity between pairs of policies. And the matrices should be full rank for effective exploration. If the kernel matrix is differentiable, we can make the behaviors of agents repel each other by maximizing probabilities characterized by the determinant. 

Determinants show great potential in modeling diversity. But in the face of real-world problems, there are still many obstacles that need to be overcome to improve population diversity. A minor problem is the high computational complexity of the determinants. The determinant may still be zero when exploitation occurs in population although it comes from a PSD matrix. A more urgent problem is to find a balance between reward and diversity. The mainstream algorithms~\cite{dvd,edocs} regard the adaptive exploration as an online learning problem, and reduce the manual process through Multi-armed Bandits algorithms (MABs)~\cite{ts,ucb}. Unfortunately, the so-called equilibrium of reward and diversity may not exist due to the complex environmental dynamics or the non-stationary distribution of reward signals for MABs in the training process. the conflict between reward and diversity makes the MABs unable to eliminate the impact of exploration on performance. 

In this paper, we mainly focus on small-scale population-based RL and address the above problems. We consider several existing probability measures as kernels to characterize the similarities between pairwise stochastic policies and construct the matrices of DPPs. To satisfy the restrictive constraint on Cholesky decomposition, we advocate the use of a regularization to improve population diversity so that the optimization can also be performed when the original matrix is singular, which is common in PBT. To avoid the conflict between reward and diversity, we introduce an archive and an auxiliary phase into the PBT framework. In our proposed Phasic Diversity Optimization (PDO) algorithm, the archive can be a MAP-Elites\cite{me} grid or a fitness priority queue without a \textbf{Behavior Descriptor} (BD). In the auxiliary phase, we diversify the potential agents from the archive to fill the blank area of the grid or eliminate the worst agents in the archive. To demonstrate the practicality of the proposed algorithm, we introduce the task for a dogfight agent, considering modeling the problem as a Markov Decision Process (MDP) for end-to-end training and evaluation. Experiments on different tasks show that PDO can diversify agents without harming reward.

\section{RELATED WORK}
\textbf{Diversity RL.} Previous works\cite{diayn,rr,trajdiv,mep} have fully discussed the importance of population diversity from different perspectives. DIAYN\cite{diayn} updates the skill of agents to visit diverse states. Reward-Randomized PG\cite{rr} perturbs the reward space of the original environment. TrajeDi\cite{trajdiv} and MEP\cite{mep} generalize the JSD and generates diverse policies through trajectory-based objectives. Diversity-Inducing PG~\cite{dipg} and Reward-Switching PG~\cite{rspo} obtain the policies of different behaviors in incremental ways. The above methods are proved to be effective in where diversity can be distinguished by different states, reward functions, and action distributions.

\textbf{High-performance diversity.} A common way to encourage exploration is to maximize an entropy regularization\cite{a3c,sac} which increases the variance of action distribution and are not directly related to the improvement of performance with increasing the complexity of convergence. The idea of Diversity-Driven Exploration Strategy (Div,~\cite{divdri}) shares a common intuition with the P3S\cite{p3s}, where both exploitation and exploration may improve performance. DvD\cite{dvd} and EDO-CS\cite{edocs}, as typical applications of adaptive exploration, using MABs, may still cause long-term performance damage due to complex reasons. The parameters of the above methods need to be fine-tuned.

\textbf{Quality-Diversity (QD) RL.} QD is a hot topic in Evolutionary Computation, in which MAP-Elites\cite{me} as the typical algorithm using BDs to map agents into the archive. Different cells represent different behaviors, and each cell stores one agent with the best fitness. Compared with RL, the Genetic Algorithm can more effectively diversify agents but have less sample efficiency. Therefore, algorithms such as QD-PG~\cite{qd-rl2} and PGA-MAP-Elites~\cite{pga-me} combining QD and TD3\cite{td3} are becoming mainstream. In contrast, our PDO uses joint optimization and can deploy any on-policy algorithm.

\textbf{Air Combat Based on Deep RL.} Recently, Deep RL has been applied to autonomously perform air combat \cite{air2,air3,air4,air5,air6}. \cite{air2} trains an agent in a custom three-dimensional environment that takes actions from a set of 15 discrete maneuvers and achieves victory over human opponents. \cite{air3} reaches its goal in an optimal path using TD3\cite{td3} + HER\cite{air7}. \cite{air4} presents a state stacking method for noisy RL environments as a noise reduction. And \cite{air5,air6} apply RL methods to air combat, but not take into account the population diversity.




\section{PRELIMINARIES}

\subsection{Stochastic Policy Gradient}
We consider RL in a MDP denoted as a tuple $(\mathcal{S},\mathcal{A},P,R, \gamma )$, where
$\mathcal{S}$ is the state space,
$\mathcal{A}$ is the action space,
$P:\mathcal{S} \times \mathcal{A} \times \mathcal{S} \to [0,1]$ is the state transition probability,
$R:\mathcal{S} \times \mathcal{A} \to \mathbb{R}$ is the reward function and
$\gamma \in [0,1]$ is the discount factor.
A stochastic policy $\pi_\theta:\mathcal{S} \times \mathcal{A} \to [0,1]$, parameterized by a vector $\theta$, assigns a probability of an action given a state. The policy's goal is to maximize the accumulated reward by maximizing the expected discounted return:
\begin{equation}
\label{eq:PG}
    J_\pi(\theta) = \mathbb{E}_{\tau \sim \pi_\theta} \left[\sum^{T-1}_{t=0}{\gamma^t r_t}\right]
\end{equation}
where trajectory $\tau (s_0,a_0,s_1,a_1,...,s_{T-1},a_{T-1})$ is generated by sampling actions according to the policy $a \sim \pi_\theta(a_t|s_t)$ and states according to the dynamics $s_{t+1} \sim P(s_{t+1}|s_t,a_t)$.

\subsection{Diversity via Determinants}

Determinantal point processes are probabilistic models of repulsion where distributions of diverse sets are characterized by determinants. From a geometric point of view, the determinant measures the directed volume of the vectors in the hyperplane space.

The Diversity via Determinants (DvD,~\cite{dvd}) algorithm maximizes the volume to boost exploration. Formally, the population diversity parameterized by $\Theta = \{ \theta_i\}_{i=1}^M$ is defined as:
\begin{equation}
\label{eq:div}
    Div(\Theta):={\det(K(\phi(\theta_{i}),\phi(\theta_{j}))_{i,j=1}^M)}
\end{equation}
where $l$ is the dimension of embedding, $K:\mathbb{R}^l \times \mathbb{R}^l \to [0,1]$ is a similarity kernel function, and $\phi: \theta \to \mathbb{R}^l$ is the policy embedding~\cite{bem}. If the entries of the kernel matrix are measurements of similarity between pairs of policies, the similar policies are unlikely to appear together. And DvD encourages diversity by introducing an explicit diversity criterion:
\begin{equation}
\label{eq:DvD}
    J(\Theta) = (1-\lambda)\sum_{i=1}^M{J_\pi(\theta_i)}+\lambda Div(\Theta)
\end{equation}
where $\lambda \in [0,1]$ is the trade-off between reward and diversity. The $\lambda$ is selected by MABs, which maintain the posterior distribution of cumulative reward over arms. Specifically, each arm represents a coefficient $\lambda$, and these arms can be selected by Thompson sampling (TS,~\cite{ts}) or Upper Confidence Bound (UCB,~\cite{ucb}) algorithm to maximize reward in limited sampling times. The reward signal for MABs comes from a Bernoulli distribution of whether the performance of the best agent has been improved after optimization with corresponding coefficients. Finally, the arm with the higher reward will be selected more times and helps improve population performance.

\section{PHASIC DIVERSITY OPTIMIZATION ALGORITHM}
\subsection{Two-phase Opimization}

\begin{algorithm}[t]
\caption{Phasic Diversity Optimization}
\label{alg:pdo}
\begin{algorithmic}
    \STATE Initialize population $\Theta = \{ \theta_j \}_{j=1}^M$, archive $\mathbb{A} = \emptyset$
    \FOR{$j = 1$ \textbf{to} $M$}
        \STATE $\mathbf{add\_to\_archive}(\theta_j,\mathbb{A})$
    \ENDFOR
    \FOR{$t = 1$ \textbf{to} $T$}
        \STATE Initialize empty buffer $\mathbb{B}$ 
        \FOR{$j = 1$ \textbf{to} $M$}
            \STATE // Optimize reward in parallel learners
            \STATE Perform rollouts $\mathbb{B}_j$ under policy $\pi_{\theta_j}$
            \STATE Maximize $J_\pi(\theta_j)$ w.r.t. $\theta_j$, on all transitions in $\mathbb{B}_j$
            \STATE Add $\mathbb{B}_j$ to buffer $\mathbb{B}$
            \STATE $\mathbf{add\_to\_archive}(\theta_j,\mathbb{A})$
        \ENDFOR
        \STATE Exploit $\mathbb{A}$ periodically
        \STATE // Optimize diversity in one learner
        \STATE Sample $M$ solutions $\hat{\Theta} = \{ \hat{\theta}_j \}_{j=1}^M$ in $\mathbb{A}$
        \STATE Maximize $J_D(\hat{\Theta})$ w.r.t. $\hat{\Theta}$, on sampled states in $\mathbb{B}$
        \FOR{$j = 1$ \textbf{to} $M$}
            \STATE $\mathbf{add\_to\_archive}(\hat{\theta}_j,\mathbb{A})$
        \ENDFOR
    \ENDFOR
\end{algorithmic}
\end{algorithm}

Several motivations drive us to propose the Phasic Diversity Optimization (PDO) algorithm. First, if coefficients come from an inappropriate space, it will affect the performance of the agent for a long time even if MABs abandon pulling these arms during training. Moreover, MABs need a stationary model of reward distribution, which is non-trivial because of the fluctuating returns, skills learned by agents, and even replication dynamics. Second, diversity may conflict with reward. We want to avoid multi-objective optimization, which may both lose quality and fail to diversify. Third, if the diversity criterion for multi-objective optimization is joint loss, additional gradient transfer overhead is required between processes. The serialized training process grows the wall-clock time in practice.

The idea of PDO is simple. It separates the training into two phases. In the first phase, we perform policy iterations and decide whether to store them in the archive based on the evaluation results. In the second (auxiliary) phase, we elect the best $M$ agents from the archive, improve their population diversity via determinants, and then evaluate them to decide whether to store them in the archive. The pseudo-code is given in Algorithm \ref{alg:pdo}. Note that the two phases can run in parallel.

The PDO algorithm improves population diversity in a way that is not harmful to reward while using as few learners as possible to diversify agents. It prevents performance degradation by archive backups. If the agents generated in the auxiliary phase are not diverse enough, they will only be compared with their previous selves and will be permanently discarded if they get worse. If the archive has a Behavior Descriptor (BD, denoted $\boldsymbol{b}$), the agents may be mapped to another cell because of repulsion and compare performance with the elite. The motivation to use BD is weakened by PDO, and if we are not experts in behavior modeling, we can use the reward-priority queues as the storage media for agents, which can be easily implemented with heaps.

Each iteration of the PDO performs two reads of the archive. The first read occurs at the exploitation of the archive, where we uniformly sample an agent from the archive to eliminate the current worst agent, with the aim of selecting targets for the policy iterations. The second read occurs to select the top $M$ agents with the aim of selecting target for the diversity iterations. We can also use the clustering-based selection\cite{edocs} algorithm to select $M$ agents for the auxiliary phase, which is different from their original motivation.

\subsection{Learning to Score Population Diversity}

From the perspective of probability theory, we propose a differentiable method to estimate the similarity between two stochastic policy distributions and form the PSD matrix $\boldsymbol{K}$.

Let $D(P,Q):\mathcal{A} \times \mathcal{A} \to \mathbb{R}$ denotes the distance metrics of any two stochastic distributions $P$ and $Q$. The distance between two stochastic policies $\pi$ and $\pi'$ defined as:
\begin{equation}
\label{eq:sim}
    d_D(\pi,\pi'):= \int_{\mathcal{S}} {D(\pi(\cdot|s),\pi'(\cdot|s)) \mathrm{d} s}
\end{equation}

\begin{definition} [Differentiable Similarity Estimation] The similarity between two stochastic policies $\pi$ and $\pi'$ can be estimated by the approximate kernel function $K$:
\begin{equation}
\label{eq:dae}
    K_{f}(\pi,\pi'):= \mathbb{E}_{s}[f(D(\pi(\cdot|s),\pi'(\cdot|s)))]
\end{equation}
where $f:\mathbb{R} \to [0,1]$ is a differentiable function.
\end{definition}

The Differentiable Similarity Estimation (DSE) allows us to sample the visited states from trajectories of mixed average policies, and estimate the similarity by finite sampling instead of the intractable integral in \eqref{eq:sim}. We can easily map the value to $[0,1]$ by the function $f$ if $D$ is a symmetric and bounded metric. Then the policy can be updated by applying the chain rule to \eqref{eq:dae} w.r.t. the parameter $\theta$:
\begin{equation}
    \nabla_{\theta} K = (\nabla_f K_f)(\nabla_D f)(\nabla_{\theta} D)
\end{equation}

In discrete action spaces, a symmetric and bounded distance $D$ between two probability measures $P$ and $Q$ and its corresponding function $f$ can be the \textbf{Jensen–Shannon divergence (JSD)}: $D_{JS}(P\|Q)=\frac{1}{2}D_{KL}(P\|\frac{P+Q}{2})+\frac{1}{2}D_{KL}(Q\|\frac{P+Q}{2})$ and $f_{JS}(d)=1-\frac{d}{\ln{2}}$, where $D_{KL}(P\|Q)=\sum_{x}{P(x)\log(\frac{P(x)}{Q(x)})}$.

However, JSD has no closed-form solution for two Gaussian distributions. In continuous action spaces, we can use the Euclidean norm as the distance function, defined by the \textbf{$p$-Wasserstein distance (WDs)}~\cite{wgan}:
\begin{equation}
\label{eq:wd}
    W_p(\mu,\nu):=\left( \inf_{\omega \in \Omega(\mu,\nu)} \int_{\mathcal{A} \times \mathcal{A}} {\|x-y\|^p\mathrm{d}\omega(x,y)} \right)^{1/p}
\end{equation}
where $\Omega(\mu,\nu)$ is the set of all couplings (joint distributions) between $\mu$ and $\nu$ on $\mathcal{A}$.

Equation \eqref{eq:wd} is inspired by the optimal transport problem (OTP) where we want to transform $\mu$ from $\nu$ with minimum cost. In the rest of the paper, we only consider the $2$-Wasserstein distance because the closed-form~\cite{wd} formula is available when $\mu$ and $\nu$ are multi-variate Gaussian distributions. If $\mu \sim N(m_1,\Sigma_1)$ and $\nu \sim N(m_2,\Sigma_2)$, we have:
\begin{equation}
\label{eq:wd2}
    W_2^2(\mu,\nu)=\|m_1-m_2\|^2+ \mathrm{tr}[\Sigma_1+\Sigma_2-2(\Sigma_1^{\frac{1}{2}} \Sigma_2\Sigma_1^{\frac{1}{2}})^\frac{1}{2}]
\end{equation}

In particular, if the $\mu$ and $\nu$ are uncorrelated multivariate normal distributions, the covariance matrices $\Sigma_1$ and $\Sigma_2$ are diagonal and \eqref{eq:wd2} can be further simplified as:
\begin{equation}
\label{eq:wd2_u}
    W_2^2(\mu,\nu)=\|m_1-m_2\|^2+ \|\Sigma_1^{\frac{1}{2}}-\Sigma_2^{\frac{1}{2}}\|_F^2
\end{equation}

It only measures the distance between the mean values of two distributions if we ignore the second term in \eqref{eq:wd2_u}. And $f$ can be the Radial Basis Function (RBF) kernel with length-scale $\sigma$:
\begin{equation}
    f_{W}(d)=\exp (-\frac{d^2}{2 \sigma^2})
\end{equation}
which is the same as the kernel function for the deterministic action embedding. Note that the result of WDs is highly correlated with the norm of $d$, so we use variance normalization to eliminate hyperparameter $\sigma^2$.

\subsection{Surrogate Determinant Regularization}

We introduce a joint diversity loss based on determinants. Since the entries of the matrix are computed by differentiable approximate kernel function, we can compute the gradient of the determinant directly by automatic differentiation.

\begin{lemma}
\label{theorem:detgrad}
The gradient of the determinant of positive definite matrix $\boldsymbol{K}$ w.r.t. $\Theta$ equals:
\begin{equation}
\label{eq:detgrad}
    \nabla_{\Theta} \det(\boldsymbol{K}) = \det(\boldsymbol{K}) \boldsymbol{K}^{-1} (\nabla_{\Theta} \boldsymbol{K})
\end{equation}
\end{lemma}

\begin{theorem}
\label{theorem:cholesky}
Consider a positive semidefinite matrix $\boldsymbol{A}$ with $\det \boldsymbol{A}>0$, $\boldsymbol{A}$ is positive definite, and the Cholesky decomposition of $\boldsymbol{A}$ is available as: $\boldsymbol{A}=\boldsymbol{L} \boldsymbol{L}^\mathrm{T}$ where $\boldsymbol{L}$ is an invertible lower (or upper) triangular matrix.
\end{theorem}

Note, due to the existence of PBT's exploitation mechanism, the network weights of a policy may be copied by another. Consequently, some two rows (or columns) in the matrix are linearly correlated, the determinant is zero, and the $\boldsymbol{K}^{-1}$ in \eqref{eq:detgrad} may encounter numerical instability (e.g., maximum likelihood estimation). In this case, $\boldsymbol{K}$ will degenerate into a PSD matrix which makes the decomposition impossible (or the gradient is zero). To meet the conditions in Theorem \ref{theorem:cholesky}, we use surrogate matrix $\tilde{\boldsymbol{K}}$:
\begin{equation}
    \tilde{\boldsymbol{K}}=\beta\boldsymbol{K}+(1-\beta)\boldsymbol{I}
\end{equation}
where $\beta \in (0,1)$ is the smoothing parameter and $\boldsymbol{I}$ is identity matrix.
\begin{lemma}
\label{lemma:pd}
$\det(\tilde{\boldsymbol{K}})>=(1-\beta+M\beta)(1-\beta)^{M-1}>0$
\end{lemma}

The surrogate determinant does not change the repulsive of the original one. Lemma \ref{lemma:pd} implies that if all the pairs of different policies in the population are not perfectly similar, and the kernel matrix is \textbf{positive definite} the condition of Theorem \ref{theorem:cholesky} is satisfied. Therefore, we derive the determinant from the principal diagonal of the lower triangular matrix $\boldsymbol{L}$ via Cholesky decomposition.
Then $\boldsymbol{K}$ is replaced by a positive definite matrix $\tilde{\boldsymbol{K}}$, and the objective in the auxiliary phase is given by:
\begin{equation}
\label{eq:pdo2}
    J_D(\Theta)=\det(\tilde{\boldsymbol{K}})
\end{equation}
The gradient of objective in \eqref{eq:pdo2} w.r.t. parameter $\theta_i \in \Theta$ is given by:
\begin{equation}
    \nabla_{\theta_i} J_D(\Theta) =  (\nabla_{\theta_i}\Theta) \nabla_\Theta \det(\tilde{\boldsymbol{K}})
\end{equation}

\begin{figure*}[t]
    \centering
    \includegraphics[width=\textwidth]{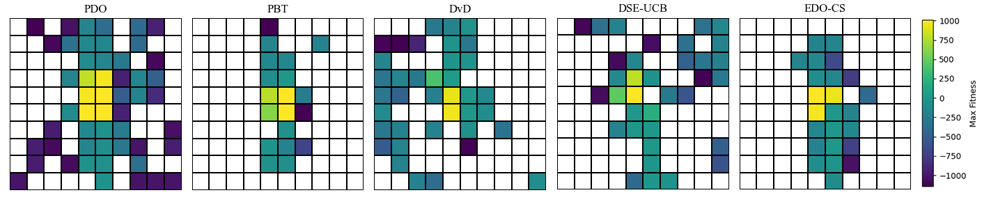}
    \caption{The archives with 2-dimensional BDs found by different algorithms in the dogfight task from the same environmental seed. Increasing the value of Elevator from top to bottom and Roll from left to right. The color bar on the right shows the different learning levels of the agent.}
    \label{fig:me}
\end{figure*}

\section{SIMULATION RESULTS}

\begin{figure}[ht]
    \centering
    \includegraphics[width=0.5\textwidth]{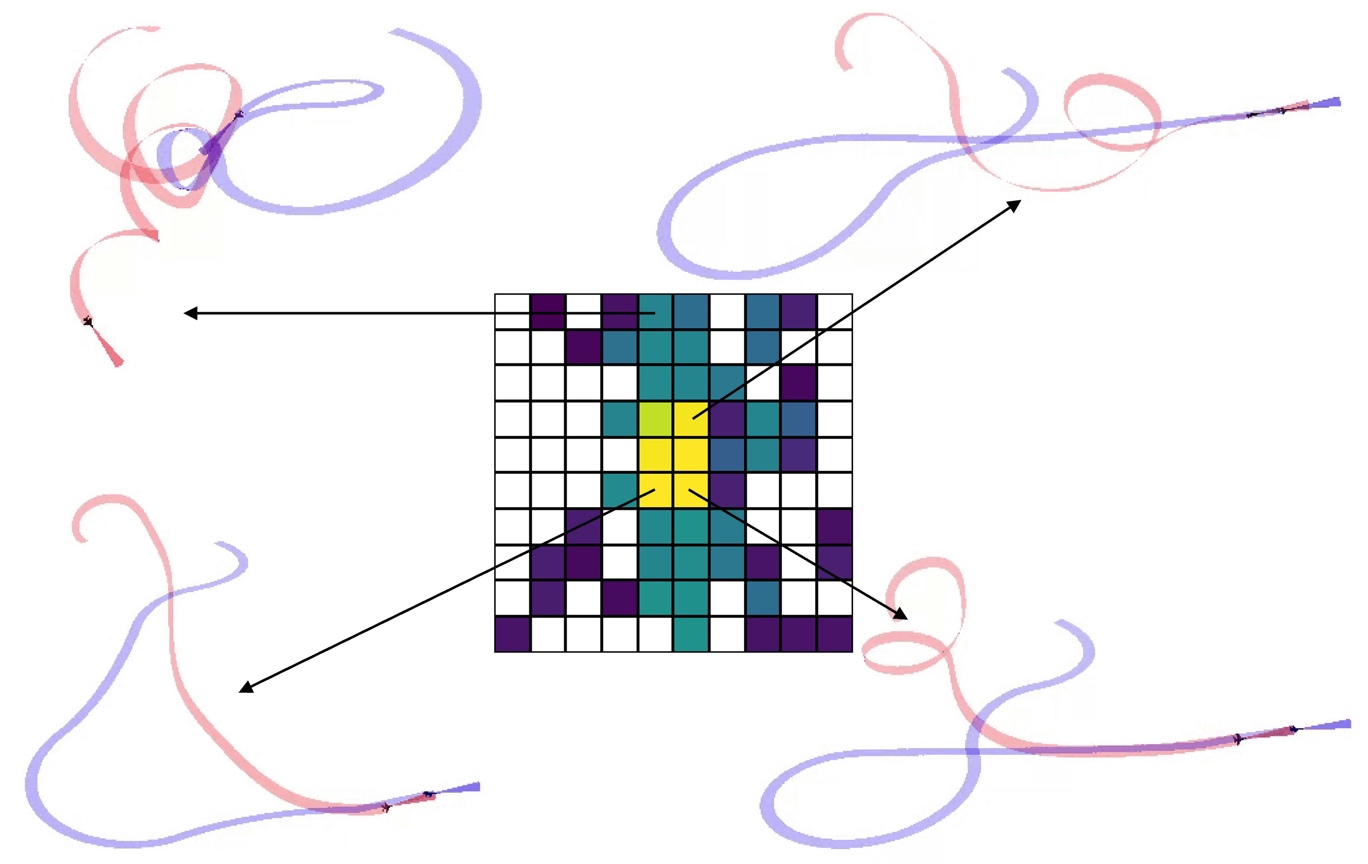}
    \caption{The trajectories from agents in different cells.}
    \label{fig:df}
\end{figure}

\subsection{2-Agent Dogfight Environment}
\label{exp:df}
In this section, we present a competition scenario that consists of 2 adversarial agents and a simplified aerodynamic three-dimensional world. An agent needs control of 4 volumes (Brake and Throttle, Elevator, Roll, and Rudder) to lock onto the target while evading its lock. The state space contains the position, velocity, and posture information of both agents. We set expert policy for the opponent, which means that the agent needs to learn evasive maneuvers and precision tracking skills in a highly dynamic environment. 

We set $1$, $-1,$ and $-1000$ rewards for locking, being locked, and out of bounds respectively. Each episode terminates at: (1) the number of steps exceeds 3000. (2) Any agent is out of bounds. (3) Any agent is locked for more than 1000 steps. Considering the lock requires the target to be within the agent's line of sight (a 10-degree cone and within 1 km) which is a harsh condition, we use a dense reward at every step during the learning process (exclude from evaluation) for all algorithms. The expert is programmed to control the elevator and rudder in order to minimize the antenna train angle and the distance from the red agent. It decelerate and apply brakes when the aspect angle is below 30 degrees and the distance is less than 3 km. The above action takes a definite value of 0.9, and the remaining values are uniformly sampled from $[-0.1,0.1]$ to improve randomness.
We consider the following techniques:
\begin{itemize}
    \item \textbf{Proximal Policy Optimization (PPO, \cite{ppo})}. The base on-policy algorithm with one learner.
    \item \textbf{Population Based Training of Neural Networks (PBT, \cite{pbt})}. It removes the auxiliary stage of the PDO algorithm, which involves parallel learners and periodic exploitation through a performance-based priority queue. This is considered as one of the ablation settings for the PDO algorithm.
    \item \textbf{PDO}. The two phases are reward optimization and diversity optimization respectively. We perform several diversity iterations in each auxiliary phase.
    \item \textbf{DvD}\cite{dvd}. It adds $\lambda \det(\boldsymbol{K})$ to the objective, where trade-off $\lambda$ is selected by Thompson sampling\cite{ts}.
    \item \textbf{DSE-UCB}. Similar to DvD, but the trade-off $\lambda$ is selected by the Upper Confidence Bound\cite{ucb}.
    \item \textbf{EDO-CS}. Similar to DSE-UCB, but it selects the agents for policy iterations through clustering-based selection\cite{edocs}.
\end{itemize}

We employ PPO\cite{ppo} and use cumulative rewards of the 10-episodes-average to evaluate agents after 25 policy iterations for all PPO variants. Each learner trains 3e6 steps. We set the population size $M=5$ and run with 5 random seeds for all Populations-Based methods, and provide MAP-Elites archives for the PDO, PBT\cite{pbt}, DvD\cite{dvd}, and DSE-UCB\cite{ucb} algorithms.  They uniformly sample the solution from the archives and replace the worst performing agent with model weights, optimizer, rewards and state normalization information after 6e5 steps. In DSE\cite{ucb}, the sample size is 256, and we use the $2$-Wasserstein distance and RBF kernel without the Frobenius norm of the covariance matrix because we use \textbf{deterministic} evaluation. We set $\{0.0, 0.5\}$ for MABs that are the same as those in paper\cite{dvd,edocs}. The number of diversity iterations is 20 in the auxiliary phase. The BD $\boldsymbol{b}$ is defined as the average values of the Elevator and Roll controlled by the agent, and the archive size is (10, 10). We visualize the archives from one of the seeds, as shown in Fig. \ref{fig:me}.

\begin{table}[t]
    \centering
    \caption{Results for the compared algorithms in the dogfight task.}
    \label{tab:baselines}
   \begin{tabular}{c|ccccc}
    \hline
     & \multicolumn{5}{c}{$M=5$}  \\
    \cline{2-6} 
    & PBT & DvD & DSE-UCB & EDO-CS & PDO \\
    \hline
    \textit{Max Fitness} & 968 & 575  & 634 & 682 & \textbf{972}\\
    \textit{QD-score} & 21032 & 30043 & 35724 & 23731  & \textbf{39201} \\
    \textit{Coverage} & 20.2 & 33.0 & 40.0 & 23.2  & \textbf{50.8} \\ 
    \hline
    \end{tabular}
\end{table}

We can observe from Table \ref{tab:baselines}\footnote{Max Fitness refers to the cumulative rewards of best agents. QD-score refers to the total sum of rewards offset by minimum rewards across all agents in the archive. Coverage refers to the total number of agents in the archive.} that among the all compared methods, the PDO algorithm exhibits the highest maximum fitness, coverage, and quality diversity scores. In terms of the maximum fitness metric, the PDO algorithm slightly outperformed the PBT\cite{pbt} algorithm.

A more advanced inquiry would be whether the strategies learned by the PDO are diverse, despite its high coverage. Fig. \ref{fig:df} displays reproducible (under the same random seed and expert policy setting) trajectory segments of the agent piloting the red aircraft and engaging the blue target. The four red trajectories correspond to the agents from the (0,4), (3,5), (5,4), and (5,5) cells in the PDO algorithm archive is shown in Fig. \ref{fig:me}. We can observe that the agents in the bright-colored cells of the PDO algorithm-generated MAP-Elites archive have learned effective attacking strategies, while the agents in the dark-colored cells have not yet learned effective attacking strategies. Thus, the agents in different cells exhibit different flight trajectories, indicating their learned behaviors and strategies are diverse.

\subsection{Performance on MuJoCo Environments}

\begin{figure*}[t]
    \centering
    \includegraphics[width=\textwidth]{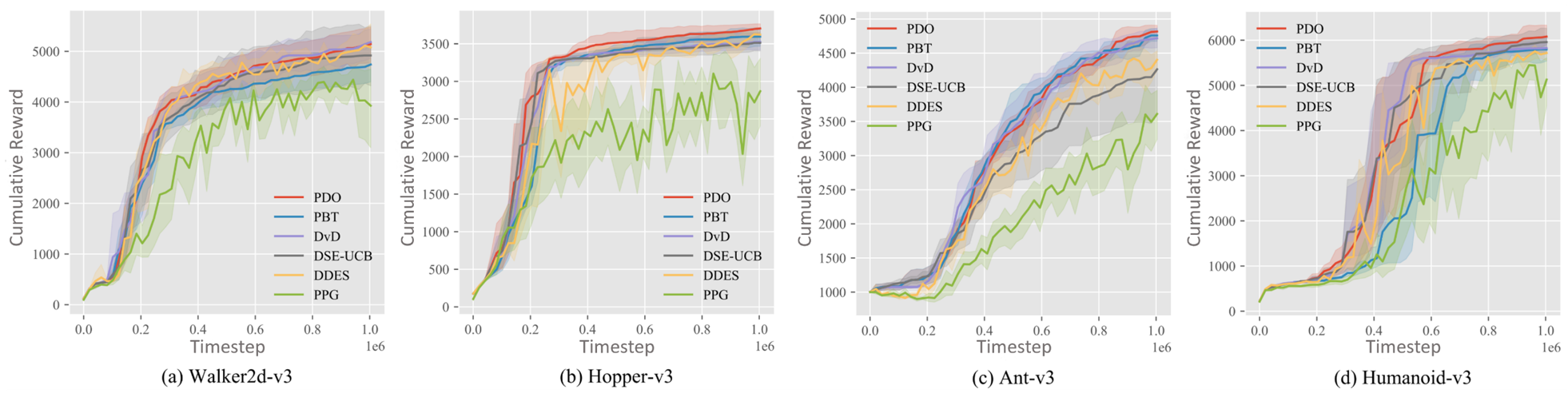}
    \caption{Average cumulative reward curves from best agent on MuJoCo tasks. The shade area represents to a single standard deviation.}
    \label{fig:mujocos}
\end{figure*}

We examine the performance of PDO on standard MuJoCo\cite{mujoco} environments from OpenAI Gym\cite{gym}. The goal of agents in MuJoCo is to move forward by controlling the joints. In order to demonstrate the advantage of PDO, we consider the following methods incorporating parallel techniques:

\begin{itemize}
    \item \textbf{Phasic Policy Gradient (PPG, \cite{ppg})}. The base on-policy algorithm with one learner.
    \item \textbf{Diversity-Driven Exploration Strategy (DDES, \cite{divdri})}. The algorithm involves maximizing a heuristic-driven driving target in the loss function.
    \item \textbf{PBT, PDO, DvD and DSE-UCB}. As described in \ref{exp:df}.
\end{itemize}

We set the same hyperparameters and PPG for all parallel learners. For all tasks, we strive to make minimal changes to PPG's hyper-parameters compared to PPO's~\cite{ppo} paper, and we only reduced the number of data usage of actors. We use $M=5$ learners with 5 random seeds for all Populations-Based methods. We use the 10-episodes-average cumulative rewards to evaluate agents after 10 policy iterations. The capacity of the priority queue is 10. The epoch of diversity iterations is 5 in the auxiliary phase. The other hyper-parameters are the same as described in \ref{exp:df}.

Fig. \ref{fig:mujocos} shows the results of different variants of PPG algorithm on four MuJoCo tasks. We can observe that the population-based algorithm demonstrates significantly better performance compared to the baseline the PPG algorithm. The final cumulative reward of the PDO algorithm surpasses other diversity-enhanced RL techniques in the Ant-v3, Hopper-v3, and Humanoid-v3 environments, while approaching the performance of the best algorithm in the Walker-v3 environment.

\begin{figure}[t]
    \centering
    \includegraphics[width=0.5\textwidth]{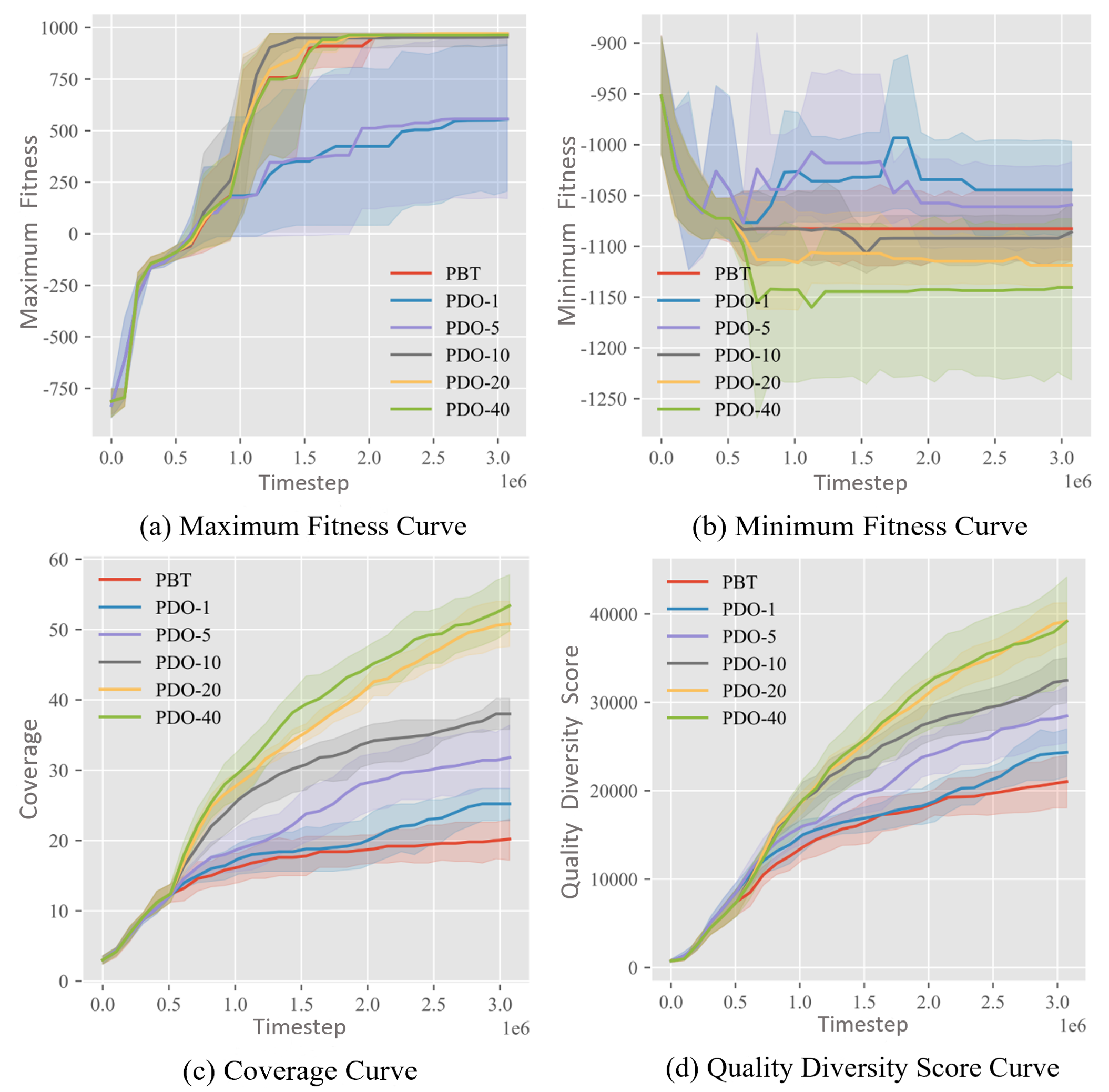}
    \caption{QD metrics curves on dogfight from different diversity epochs}
    \label{fig:humanoid}
\end{figure}

\subsection{Ablation Studies}

We conduct ablation experiments on PDO and performed sensitivity analysis on different levels of diversity during the auxiliary phase. Each experiment runs with 5 random seeds and we record the mean and standard deviation. When PDO removes the auxiliary phase, it becomes the PBT algorithm. Fig. \ref{fig:humanoid} illustrates this concept, where Subfig. \ref{fig:mujocos}(a)-(d) display the curves of maximum fitness, minimum fitness, coverage range, and quality diversity score for different levels of diversity. The curves represent the mean values, while the shaded areas indicate the standard deviation.

From the four subfigures in Fig. \ref{fig:humanoid}, we can conclude that the PDO algorithm significantly outperforms the ablated PBT algorithm in terms of coverage range and QD-score indicators of the archive. Subfig. \ref{fig:humanoid}(a) infers that the PDO algorithm does not overlook performance due to excessive diversification. Subfig. \ref{fig:humanoid}(b) implies a negative correlation between the quantity of diversity and the minimum fitness, indicating that a higher degree of diversity iterations may result in the learning of less effective strategies.

\section{CONCLUSIONS}
In this paper, we propose PDO algorithm to achieve the decoupling of reward and diversity. We extend the DvD algorithm to a more general form compatible with PBT. We practically apply the proposed framework to the well-known policy gradient algorithms\cite{ppo,ppg}. Experiments on our dogfight task and MuJoCo show that our simple modification on PBT is effective and brings further improvement to the performance.





\section*{ACKNOWLEDGMENT}
This work was supported in part by National Key Research and Development Program of China (2022ZD0210500/2021ZD0112400), the National Natural Science Foundation of China under Grants 62102058/62272081/61972067/62332019, the Distinguished Young Scholars Funding of Dalian (No. 2022RJ01), and the open funding of State Key Laboratory of Structural Analysis for Industrial Equipment.

\bibliographystyle{unsrt}
\bibliography{ref}

\begin{thebibliography}{10}

\bibitem{pbt}
Max Jaderberg, Valentin Dalibard, Simon Osindero, Wojciech~M. Czarnecki, Jeff Donahue, Ali Razavi, Oriol Vinyals, Tim Green, Iain Dunning, Karen Simonyan, Chrisantha Fernando, and Koray Kavukcuoglu.
\newblock Population based training of neural networks.
\newblock {\em arXiv preprint arXiv:1711.09846}, 2017.

\bibitem{pga-me}
Olle Nilsson and Antoine Cully.
\newblock Policy gradient assisted map-elites.
\newblock In Francisco Chicano and Krzysztof Krawiec, editors, {\em {GECCO} '21: Genetic and Evolutionary Computation Conference, Lille, France, July 10-14, 2021}, pages 866--875. {ACM}, 2021.

\bibitem{divauto}
Yaodong Yang, Jun Luo, Ying Wen, Oliver Slumbers, Daniel Graves, Haitham Bou{-}Ammar, Jun Wang, and Matthew~E. Taylor.
\newblock Diverse auto-curriculum is critical for successful real-world multiagent learning systems.
\newblock In {\em {AAMAS} '21}, pages 51--56, 2021.

\bibitem{neuro}
F{\'{e}}lix Chalumeau, Raphael Boige, Bryan Lim, Valentin Mac{\'{e}}, Maxime Allard, Arthur Flajolet, Antoine Cully, and Thomas Pierrot.
\newblock Neuroevolution is a competitive alternative to reinforcement learning for skill discovery.
\newblock {\em arXiv preprint arXiv:2210.03516}, 2022.

\bibitem{dpp}
Alex Kulesza and Ben Taskar.
\newblock Determinantal point processes for machine learning.
\newblock {\em Found. Trends Mach. Learn.}, 5(2-3):123--286, 2012.

\bibitem{dvd}
Jack Parker-Holder, Aldo Pacchiano, Krzysztof~M Choromanski, and Stephen~J Roberts.
\newblock Effective diversity in population based reinforcement learning.
\newblock In {\em Advances in Neural Information Processing Systems}, volume~33, pages 18050--18062, 2020.

\bibitem{edocs}
Yutong Wang, Ke~Xue, and Chao Qian.
\newblock Evolutionary diversity optimization with clustering-based selection for reinforcement learning.
\newblock In {\em The Tenth International Conference on Learning Representations, {ICLR} 2022, Virtual Event, April 25-29, 2022}. OpenReview.net, 2022.

\bibitem{ts}
William~R Thompson.
\newblock On the likelihood that one unknown probability exceeds another in view of the evidence of two samples.
\newblock {\em Biometrika}, 25(3-4):285--294, 1933.

\bibitem{ucb}
Peter Auer, Nicol{\`{o}} Cesa{-}Bianchi, and Paul Fischer.
\newblock Finite-time analysis of the multiarmed bandit problem.
\newblock {\em Mach. Learn.}, 47(2-3):235--256, 2002.

\bibitem{me}
Jean{-}Baptiste Mouret and Jeff Clune.
\newblock Illuminating search spaces by mapping elites.
\newblock {\em arXiv preprint arXiv:1504.04909}, 2015.

\bibitem{diayn}
Benjamin Eysenbach, Abhishek Gupta, Julian Ibarz, and Sergey Levine.
\newblock Diversity is all you need: Learning skills without a reward function.
\newblock In {\em 7th International Conference on Learning Representations}, 2019.

\bibitem{rr}
Zhenggang Tang, Chao Yu, Boyuan Chen, Huazhe Xu, Xiaolong Wang, Fei Fang, Simon~Shaolei Du, Yu~Wang, and Yi~Wu.
\newblock Discovering diverse multi-agent strategic behavior via reward randomization.
\newblock In {\em 9th International Conference on Learning Representations}, 2021.

\bibitem{trajdiv}
Andrei Lupu, Brandon Cui, Hengyuan Hu, and Jakob~N. Foerster.
\newblock Trajectory diversity for zero-shot coordination.
\newblock In {\em Proceedings of the 38th International Conference on Machine Learning}, volume 139, pages 7204--7213. {PMLR}, 2021.

\bibitem{mep}
Rui Zhao, Jinming Song, Haifeng Hu, Yang Gao, Yi~Wu, Zhongqian Sun, and Yang Wei.
\newblock Maximum entropy population based training for zero-shot human-ai coordination.
\newblock {\em arXiv preprint arXiv:2112.11701}, 2021.

\bibitem{dipg}
Muhammad~A. Masood and Finale Doshi{-}Velez.
\newblock Diversity-inducing policy gradient: Using maximum mean discrepancy to find a set of diverse policies.
\newblock In Sarit Kraus, editor, {\em Proceedings of the Twenty-Eighth International Joint Conference on Artificial Intelligence, {IJCAI} 2019}, pages 5923--5929. ijcai.org, 2019.

\bibitem{rspo}
Zihan Zhou, Wei Fu, Bingliang Zhang, and Yi~Wu.
\newblock Continuously discovering novel strategies via reward-switching policy optimization.
\newblock In {\em International Conference on Learning Representations}, 2022.

\bibitem{a3c}
Volodymyr Mnih, Adri{\`{a}}~Puigdom{\`{e}}nech Badia, Mehdi Mirza, Alex Graves, Timothy~P. Lillicrap, Tim Harley, David Silver, and Koray Kavukcuoglu.
\newblock Asynchronous methods for deep reinforcement learning.
\newblock In {\em Proceedings of the 33nd International Conference on Machine Learning}, volume~48, pages 1928--1937, 2016.

\bibitem{sac}
Tuomas Haarnoja, Aurick Zhou, Pieter Abbeel, and Sergey Levine.
\newblock Soft actor-critic: Off-policy maximum entropy deep reinforcement learning with a stochastic actor.
\newblock In {\em Proceedings of the 35th International Conference on Machine Learning}, volume~80, pages 1856--1865. {PMLR}, 2018.

\bibitem{divdri}
Zhang{-}Wei Hong, Tzu{-}Yun Shann, Shih{-}Yang Su, Yi{-}Hsiang Chang, Tsu{-}Jui Fu, and Chun{-}Yi Lee.
\newblock Diversity-driven exploration strategy for deep reinforcement learning.
\newblock In {\em Advances in Neural Information Processing Systems 31}, pages 10510--10521, 2018.

\bibitem{p3s}
Whiyoung Jung, Giseung Park, and Youngchul Sung.
\newblock Population-guided parallel policy search for reinforcement learning.
\newblock In {\em 8th International Conference on Learning Representations}, 2020.

\bibitem{qd-rl2}
Thomas Pierrot, Valentin Mac{\'{e}}, F{\'{e}}lix Chalumeau, Arthur Flajolet, Geoffrey Cideron, Karim Beguir, Antoine Cully, Olivier Sigaud, and Nicolas Perrin{-}Gilbert.
\newblock Diversity policy gradient for sample efficient quality-diversity optimization.
\newblock In Jonathan~E. Fieldsend and Markus Wagner, editors, {\em {GECCO} '22: Genetic and Evolutionary Computation Conference, Boston, Massachusetts, USA, July 9 - 13, 2022}, pages 1075--1083. {ACM}, 2022.

\bibitem{td3}
Scott Fujimoto, Herke van Hoof, and David Meger.
\newblock Addressing function approximation error in actor-critic methods.
\newblock In {\em Proceedings of the 35th International Conference on Machine Learning}, volume~80, pages 1582--1591. {PMLR}, 2018.

\bibitem{air2}
Q.~Yang, J.~Zhang, G.~Shi, J.~Hu, and Y.~Wu.
\newblock Maneuver decision of uav in short-range air combat based on deep reinforcement learning.
\newblock {\em IEEE Access}, 8, 2019.

\bibitem{air3}
Muhammed~Murat {\"{O}}zbek and Emre Koyuncu.
\newblock Reinforcement learning based air combat maneuver generation.
\newblock {\em CoRR}, abs/2201.05528, 2022.

\bibitem{air4}
Ahmet~Semih Tasbas, Safa~Onur Sahin, and Nazim~Kemal Ure.
\newblock Reinforcement learning based self-play and state stacking techniques for noisy air combat environment.
\newblock {\em CoRR}, abs/2303.03068, 2023.

\bibitem{air5}
Haiyin Piao, Yue Han, Hechang Chen, Xuanqi Peng, Songyuan Fan, Yang Sun, Chen Liang, Zhimin Liu, Zhixiao Sun, and Deyun Zhou.
\newblock Complex relationship graph abstraction for autonomous air combat collaboration: {A} learning and expert knowledge hybrid approach.
\newblock {\em Expert Syst. Appl.}, 215:119285, 2023.

\bibitem{air6}
Zhixiao Sun, Huahua Wu, Yandong Shi, Xiangchao Yu, Yifan Gao, Wenbin Pei, Zhen Yang, Haiyin Piao, and Yaqing Hou.
\newblock Multi-agent air combat with two-stage graph-attention communication.
\newblock {\em Neural Comput. Appl.}, 35(27):19765--19781, 2023.

\bibitem{air7}
Marcin Andrychowicz, Alex~Ray Filip~Wolski, Jonas Schneider, Rachel Fong, Peter Welinder, Bob McGrew, Josh Tobin, Pieter Abbeel, and Wojciech Zaremba.
\newblock Hindsight experience replay.
\newblock 2017.

\bibitem{bem}
Aldo Pacchiano, Jack Parker{-}Holder, Yunhao Tang, Krzysztof Choromanski, Anna Choromanska, and Michael~I. Jordan.
\newblock Learning to score behaviors for guided policy optimization.
\newblock In {\em Proceedings of the 37th International Conference on Machine Learning}, volume 119, pages 7445--7454. {PMLR}, 2020.

\bibitem{wgan}
Mart{\'{\i}}n Arjovsky, Soumith Chintala, and L{\'{e}}on Bottou.
\newblock Wasserstein {GAN}.
\newblock {\em arXiv preprint arXiv:1701.07875}, 2017.

\bibitem{wd}
Victor~M. Panaretos and Yoav Zemel.
\newblock Statistical aspects of wasserstein distances.
\newblock {\em Annual Review of Statistics ant Its Application}, 6(1), 2019.

\bibitem{ppo}
John Schulman, Filip Wolski, Prafulla Dhariwal, Alec Radford, and Oleg Klimov.
\newblock Proximal policy optimization algorithms.
\newblock {\em arXiv preprint arXiv:1707.06347}, 2017.

\bibitem{mujoco}
Emanuel Todorov, Tom Erez, and Yuval Tassa.
\newblock Mujoco: {A} physics engine for model-based control.
\newblock In {\em 2012 {IEEE/RSJ} International Conference on Intelligent Robots and Systems}, pages 5026--5033. {IEEE}, 2012.

\bibitem{gym}
Greg Brockman, Vicki Cheung, Ludwig Pettersson, Jonas Schneider, John Schulman, Jie Tang, and Wojciech Zaremba.
\newblock Openai gym.
\newblock {\em arXiv preprint arXiv:1606.01540}, 2016.

\bibitem{ppg}
Karl Cobbe, Jacob Hilton, Oleg Klimov, and John Schulman.
\newblock Phasic policy gradient.
\newblock In {\em Proceedings of the 38th International Conference on Machine Learning}, volume 139, pages 2020--2027. {PMLR}, 2021.

\end{thebibliography}

\end{document}